# Evaluation of Predictive Data Mining Algorithms in Erythemato-Squamous Disease Diagnosis.

Kwetishe Danjuma[1], Adenike Osofisan[2]

[1] Department of Computer Science, Modibbo Adama University of Technology,
Yola, Adamawa State, Nigeria

[2] Department of Computer Science, University of Ibadan,
Ibadan, Oyo State, Nigeria

**Abstract**
A lot of time is spent searching for the most performing data mining algorithms applied in clinical diagnosis. The study set out to identify the most performing predictive data mining algorithms applied in the diagnosis of Erythemato-squamous diseases. The study used Naive Bayes, Multilayer Perceptron and J48 decision tree induction to build predictive data mining models on 366 instances of Erythemato-squamous diseases datasets. Also, 10-fold cross-validation and sets of performance metrics were used to evaluate the baseline predictive performance of the classifiers. The comparative analysis shows that the Naive Bayes performed best with accuracy of 97.4%, Multilayer Perceptron came out second with accuracy of 96.6%, and J48 came out the worst with accuracy of 93.5%. The evaluation of these classifiers on clinical datasets, gave an insight into the predictive ability of different data mining algorithms applicable in clinical diagnosis especially in the diagnosis of Erythemato-squamous diseases.

## 1. Introduction

Predictive analytic is a branch of data mining concerned with the analysis of data to identify underlying trends, patterns, or relationships to predict future probabilities and trends [1]. It encompasses statistics, data mining and game theory that analyze current and historical facts to make predictions about future events of interest [2]. In predictive modelling, data is collected, a statistical model is formulated, predictions are made and the model is validated or revised as additional data becomes available [3]. Clinical data mining is based on strategic research to retrieve, analyze and interpret both qualitative and quantitative information available from medical datasets or records [4].

Predictive data mining automatically create classification model from training dataset, and apply such model to automatically predict other classes of unclassified datasets [5]. Predictive data mining deals with learning models to support clinicians in diagnostics, therapeutic, or monitoring tasks [6]. It learns from past experience and apply knowledge gained to future situations [7], by applying machines learning methods to build multivariate models from clinical data and subsequently make inferences on unknown data [8]. Machine learning model is related to the exploitation of supervised classification approaches. Prior to applying the learning model, the data is pre-processed to remove noise and ensure data mining principle is applied on real data [9]. Predictive data mining is the most common type of data mining that has the most application in business and real life, that is centered on data pre-processing, data mining and data post-processing collectively referred to as Knowledge Discovery in Databases [7]. Examples include the prediction of surgery outcome [10], breast cancer survival [11] and coronary heart disease risk [12] and [13] from variables such as age, sex, smoking and status, hypertension and various biomarkers. Erythemato-squamous disease (ESD) is a common skin disease in outpatient dermatology departments. The variants of ESD includes psoriasis, seboreic dermatitis, lichen planus, pityriasis rosea, chronic dermatitis and pityriasis rubra pilaris [14], [15], and [16]. In this paper, we set out to identify the most performing predictive data mining algorithms applied in the diagnosis of Erythemato-squamous skin disease.

## 2. Literature Review

### 2.1 Application of Predictive Data Mining in Clinical Diagnosis

In disease diagnosis, [17] compared rule based Repeated Incremental Pruning to Produce Error Reduction (RIPPER), Decision Tree (DT), Artificial Neural Networks (ANN) and Support Vector Machine (SVM) on the basis of Sensitivity, Specificity, Accuracy, Error Rate, True Positive Rate and False Positive Rate, and 10-fold cross validation to measure the unbiased estimate of these prediction models. SVM model was adjudged the best classifier for Cardiovascular Disease (CVD) diagnosis. [18], compared DT, k-Nearest

Neighbor (kNN), ANN, and classification based on clustering in the prediction of heart disease, and genetic algorithm (GA) to improve the predictive abilities of the chosen models. [19], demonstrated how to implement an evidence-based clinical expert system of a Bayesian model to detect coronary artery disease. The Bayesian was considered to have considerable advantage in dealing with several missing variables compared to logistics and linear regression models. In the diagnosis of Asthma with expert system, [20] did a comparative analysis of machine learning algorithms such as Auto-associative Memory Neural Networks (AMNN), Bayesian networks, ID3 and C4.5 and found AMNN to perform best in terms of algorithms efficiency and accuracy of disease diagnosis. In a study of Phospholipidosis, [21] used structure-activity relationships (SAR) to compare kNN, DT, SVM and artificial immune systems algorithms trained to identify drugs with Phospholipidosis potentials and SVM produced the best predictions followed by a Multilayer Perceptron artificial neural network, logistic regression, and kNN. In the diagnosis of Chronic Obstructive Pulmonary and Pneumonia diseases (COPPD), [22] compared neural networks and artificial immune systems. Also, [23] used DT, Naïve Bayes, and Neural Networks to analyzed heart disease. The neural network algorithm was found to predict heart disease with highest accuracy.

2.2 Erythemato-Squamous Diseases (ESD) Diagnosis

An Extreme learning machine and Artificial Neural Network was proposed by [24] to better identify the differential problem of Erythemato-Squamous skin diseases. A support vector machine (SVM) based on random subspace (RS) and feature selection was used to solve differential problem of Erythemato-Squamous diseases [25]. A Catfish Binary Particle Swarm Optimization (CatfishBPSO), Kernelized Support Vector Machines (KSVM), and Association Rule feature selection method was used to diagnose Erythemato-squamous diseases [26]. The AR-CatfishBPSO-KSVM model gained baseline performance accuracy of 99.09% compared to Support vector machines and association rules - multilayer perceptron (AR-MLP). A new model based on Adaptive Neuro-Fuzzy Inference System (ANFIS) was used in the detection of Erythemato-squamous diseases [16]. They claim ANFIS has some potentials in detecting the Erythemato-squamous diseases. Another diagnosis model based on Support Vector Machine (SVM) with a novel hybrid feature selection method - Improved F-score and Sequential Forward (IFSFFS) was used on five random training-test partitions of the Erythemato-squamous diseases datasets from University of California Irvine (UCI) machine learning repository database [27]. The result shows that SVM-based model with IFSFFS achieved the optimal classification accuracy. [28], used Rough-Neuro hybrid method to achieve accurate Erythemato-squamous diseases diagnosis. The method incorporates Rough sets Johnson Reducer for reduction of relevant attributes and artificial neural network Levenberg-Marquardt algorithm for the classification of the diseases. The model was claimed to have diagnose the diseases at an accuracy of 98.8%. Two-stage hybrid feature selection algorithms for diagnosing Erythemato-squamous diseases. It incorporate Support Vector Machines (SVM) as a classification tool, and the extended Sequential Forward Search (SFS), Sequential Forward Floating Search (SFFS), and Sequential Backward Floating Search (SBFS), as search strategies, and the generalized F-score (GF) to evaluate the importance of each feature. The two-stage hybrid model was claimed to have achieved better classification accuracy when compared to available algorithms for Erythemato-squamous diseases [29].

## 3. Predictive Data Mining (PDM) Algorithms to Compare

3.1 Decision Tree (DT) Algorithm

Decision tree is a predictive data mining techniques often used in clinical medicine to easily visualize, and understand resistant to noise in data. And is applicable in both regression and association data mining tasks [30] capable of handling continuous attributes, which are essential in case of medical data e.g. blood pressure, temperature, etc. It is a non-parametric supervised learning method used for classification to create models that predicts the value of a target variable by learning simple decision rules inferred from the data features [31]. Decision trees are significantly faster than neural networks with a shorter learning curve that is mainly used in the classification and prediction to represent knowledge. The instances are classified by sorting them down the tree from the root node to some leaf node. The nodes are branched based on if-then condition [32]. The variants of decision tree algorithm includes CART, ID3, C4.5, SLIQ, and SPRINT [33]. C4.5 algorithms is an extension of ID3 algorithms that uses concept of divide-and-conquer to construct a tree from a training set $S$ [34], and each path in the decision tree can be regarded as a decision rule [35]. The decision tree is built of nodes which specify conditional attributes – symptoms $S = \{s_1, s_2, ..., s_i\}$, branches which show the values of $v_{i,k}$ i.e. the $h-th$ range for $i-th$ symptom and leaves which present decisions $D = \{d_1, d_2, ..., d_k\}$ and their binary values,

$w_{dk} = \{0,1\}$. A sample decision tree is presented in Fig.1, and as a set of association rules in equation (1).

$$(S_1, v_{1,1}) \cap (S_2, v_{2,1}) \Rightarrow (d_1 = 1)$$
$$(S_1, v_{1,1}) \cap (S_2, v_{2,2}) \Rightarrow (d_1 = 0)$$
$$(S_1, v_{1,2}) \cap (S_3, v_{3,1}) \Rightarrow (d_2 = 1)$$
$$(S_1, v_{1,2}) \cap (S_3, v_{3,2}) \Rightarrow (d_2 = 0)$$
$$(S_1, v_{1,3}) \Rightarrow (d_3 = 1) \quad (1)$$

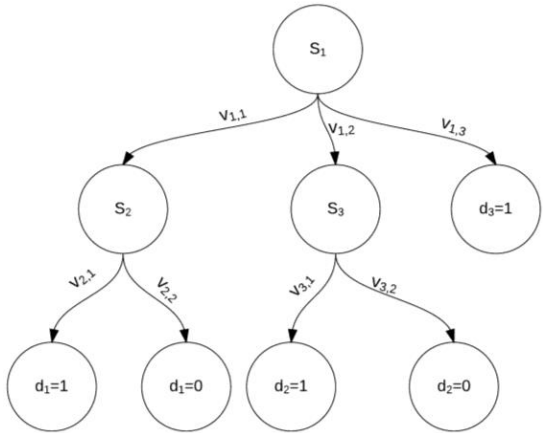

**Fig. 1** Sample decision tree applicable in clinical medicine

## 3.2 Artificial neural networks (ANN's) Algorithm

The main function of Artificial Neural Networks is prediction. Despite its complexity and difficulty in understanding the predictions, it has been successfully applied in clinical medicine especially in prediction of coronary artery disease, EEG signals processing and the development of novel antidepressants [30]. ANN is biologically inspired, highly sophisticated analytical techniques, capable of modelling extremely complex non-linear functions [36], that are modelled based on the cognitive learning process and the neurological functions of the human brain, consisting of millions of neurons interconnected by synapses [37], and capable of predicting new observations after learning from existing data. Neural networks are also capable to predict changes and events in the system after the process of learning [32]. The interconnected sets of neurons are divided into three: input, hidden, and output ones. In clinical medicine, the patient's symptoms could serve as input set $S$, and disease could serve as output set $D$ to the neural network. The hidden neuron processes the outcome of preceding layers. The attributes that are passed as input to the network form the first layer, and the direction of the network symbolizes the dataflow during the process of prediction. The process of learning in ANN is to solve a task $T$, having a set of observations and a class of functions $F$, which is to find $f^* \in F$ as the optimal solution to the task [30]. The most popular of the ANN being Multilayer Perceptron algorithm. MLP is most suitable for approximating a classification function, and consists of a set of sensory elements that make up the input layer, one or more hidden layers of processing elements, and the output layer of the processing elements [34]. The Multi-Layer Perceptron (MLP) with back-propagation (a supervised learning algorithm) is arguably the most commonly used and well-studied ANN architecture capable of learning arbitrarily complex nonlinear functions to arbitrary accuracy levels [17]. It is essentially the collection of nonlinear neurons (perceptron's) organized and connected to each other in a feed forward multi-layer structure as shown in Fig [38], defines ANN based on: a) Interconnection pattern between different layers of neurons; b) Learning process for updating the weights of the interconnection; and c) Activation function that converts a neuron's weighted input to its output activation.

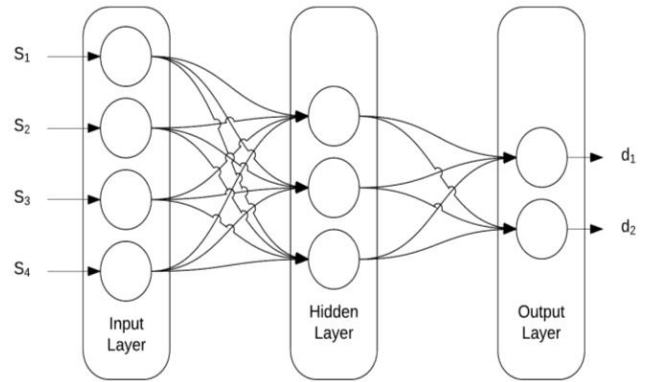

Fig. 2 Graphical depiction of ANN for clinical diagnosis for symptoms ($S_1, S_2, S_3, S_4$) like age, blood pressure, etc. and diagnosis ($d_1, d_2$)

## 3.3 Naïve Bayes Algorithm

Naïve Bayes is a Bayesian Network based on Bayes rules of simple conditional probability that estimate the likelihood of a property given the set of input data. Naïve Bayes require small amount of training data to estimate parameters such as mean and variance necessary for classification [39]. The structure of a Naïve Bayes model forms a Bayesian network of nodes with one node for each attribute. The nodes are interconnected with directed edges and form a directed acyclic graph [30] Bayesian networks uses directed acyclic graphs to model the dependencies among variables [40]. Each node in the acyclic directed graph represents a stochastic variable and arcs represent a probabilistic dependency between a node and its parents [35]. The Bayesian Network is a way of representing probabilistic relationships between variables associated with an outcome of interest. The outcome of interest could be uncertainty essential during clinical diagnosis, prediction of patient's prognosis and treatment selection. The probabilities applied in the Naïve Bayes algorithm are calculated according to the Bayes' Rule. The probability of the likelihood of some conclusion $S$, given some evidence or observation T, where $\exists$ a dependence relationship between $S$ and $T$, denoted as $P(S|T)$, can be calculated based on Eq. 1;

$$P(S|T) = \frac{P(T|S) * P(S)}{P(T)} \quad (1)$$

This conditional relationship allows an investigator to gain probability information about either $S$ or $T$ with the known outcome of the other [41].

## 4. Predictive Model Performance Evaluation Metrics

Evaluation of predictive data mining algorithms can be compared according to a number of measures. In comparing the performance of different predictive data mining algorithms to determine its predictability, some quantities that interpret the goodness of fit of a model, and error measurements must be considered [7]. Though empirical studies claimed that it is difficult to decide which metric to use for different problem, each of them has specific features that measures various aspects of the algorithms being evaluated. It is often difficult to state which metrics is the most suitable to evaluate algorithms in clinical medicine due to large weighted discrepancies that often arise between predicted and actual value or otherwise. For instance, some metrics such as true positive rate (TPR) take higher values if the algorithm gives better results compared to metric such as errors that takes lower values [30].

4.1 Accuracy, Sensitivity & Specificity (ASS)

ASS are statistical measures commonly used to explain clinical diagnostic test, and to estimate how good and consistent was the diagnostic test [42]. In clinical medicine, the ability to detect patients with sickness or exclude patients without sickness is often described by terms such as sensitivity, specificity, positive predictive value and negative predictive value. The True Positive Rate, False Positive Rate, True Negative Rate and False Negative Rate determines the predictive data mining efficiency [43]. The sensitivity metrics is also called true positive rate or positive class accuracy, while specificity is referred to as true negative rate or negative class accuracy [44]. In this study, we used three predictive model performance evaluation metrics; accuracy (Eq. (2)), sensitivity (Eq. (3)), and Specificity (Eq. (4)).

a. Accuracy-compares how close a new test value is to a value predicted by if ... then rules [45], written as in Eq.2;

$$\text{Accuracy} = \frac{TP + TN}{TP + TN + FP + FN} 100\% \quad (2)$$

b. Sensitivity-measures the ability of a test to be positive when the condition is actually present. It is often recognize as false-negative rate, recall, Type II error, β error, error of omission, or alternative hypothesis [45], written as in Eq.3;

$$\text{Sensitivity} = \frac{TP}{TP + FN} 100\% \quad (3)$$

c. Specificity-measures the ability of a test to be negative when the condition is actually not present. It is often recognize as false-positive rate, precision, Type I error, α error, error of commission, or null hypothesis [45], written as in Eq.4;

$$\text{Specificity} = \frac{TN}{TN + FP} 100\% \quad (4)$$

d. Predictive Accuracy (PA)-gives an overall evaluation. It is also known as the percentage proportion of correctly classified cases to all cases in the set. The larger the predictive accuracy the better the situation. The predictive accuracy is written as in Eq.5

$$\text{PA} = \frac{TP + TN}{TP + TN + FP + FN} 100\% \quad (5)$$

4.2 $n-$Fold cross-validation

The $n-$Fold cross-validation or rotation estimation is a model used to estimate how accurately a model will perform in practice. It involves splitting at random a sample data $R$ into complimentary $f$ mutually exclusive subsets (the folds: $R_1, R_2, ..., R_f$) for approximately equal size [11]. In order to reduce variability, the classification model is trained and tested $f$ times. Each time $(t\{1,2,...,f\})$, it is trained on all but one-fold $(Rt)$ known as training set, and tested on the remaining single fold $(Rt)$, known as validation or testing set. The cross-validation estimate of the overall accuracy is the average of the $n$ individual accuracy measures represented by Eq.6

$$CA = \frac{1}{f}\sum_{i=1}^{f} Ai \quad (6)$$

Where $CA$ stands for Cross-Validation accuracy, $f$ is the number of folds used, and $A$ is the accuracy measure of each fold [11] and [46]. Since the cross-validation accuracy depends on the random assignment of the individual cases into $n$ distinct folds, it is often stratified by creating folds that contain approximately the same proportion of predictor labels as the original dataset to generate results with lower bias and variance as compared to regular $n$-fold cross-validation [47].

4.3 Receiver Operating Characteristics (ROC)

ROC is the ratio of $FP$ and $TP$ rate (axis y). It plots the curve with $x-$ axis representing $FP$ rate and $y-$ axis representing the $TP$ rate. The ROC curve has been recommended as an appropriate measure of diagnostic accuracy by clinical epidemiologists [30]. ROC curve (AUC) is a metric of algorithms' performance. The larger the area under the curve (AUC), the better the model.

4.4 Recall, precision and F-Measure

The recall, precision and F-measure are additional parameters that could help physician determine exactly whether a patient is ill or not. Recall is the same in application as sensitivity, F-measure is the harmonic mean of both recall and precision, while specificity is the reverse of sensitivity presented by Eq.7, Eq.8 and Eq.9;

$$\text{Precision} = \frac{TP}{TP+FP} \quad (7)$$

$$\text{Recall} = \frac{TP}{TP+FN} \quad (8)$$

$$\text{F-measure} = \frac{2*precision*recall}{precision+recall} \quad (9)$$

Other performance metrics used in this study are presented in Eq.10, Eq.11, Eq.12 and Eq.13;

a. Mean absolute error

$$\frac{|p_1-a_1|+...+|p_n-a_n|}{n} \quad (10)$$

b. Root mean square error

$$\sqrt{\frac{(p_1-a_1)^2+...+(p_n-a_n)^2}{n}} \quad (11)$$

c. Relative absolute error

$$\frac{|p_1-a_1|+...+|p_n-a_n|}{|a_1-\bar{a}|+...+|a_n-\bar{a}|} \quad (12)$$

d. Root relative squared error

$$\sqrt{\frac{(p_1-\bar{a})^2+...+(p_n-a_n)^2}{|a_1-\bar{a}|^2+...+|a_n-\bar{a}|^2}} \quad (13)$$

Where $\bar{a} = \frac{1}{n}\sum_i ai$, $p$ represents predicted target values: $p_1, p_2, ..., p_n$ and $a$ represents actual value: $a_1, a_2, ..., a_n$

## 5. Empirical Results and Analysis

5.1 Erythemato-Squamous Diseases Data Analysis

In this study, we used ESD dataset from the UCI machine learning repository database. The data contains 366 number of instances, and 34 attributes. Patients were first evaluated clinically with 12 features. Afterwards, skin samples were taken for the evaluation of 22 histopathological features. The values of the histopathological features are determined by an analysis of the samples under a microscope. The family history feature with value 1 means any of these features has been observed in the family and 0 otherwise. The age feature is the patient age. Every other clinical or histopathological feature was assigned a value in the range of 0 - 3. Where, 0 means the absence of the feature, 3 means the largest amount possible, and 1, 2 means the relative intermediate values [29]. All attributes takes 3 values except the diagnosis attributes that take values in the range of 1-6, and the family-history attribute is binomial. Only the age attribute is continuous attributes with 8 missing distinguished with '?'. We removed the missing values in the age attribute because in clinical medicine, it is difficult to specify a likely range of values to replace missing values without being biased [30]. The details of the dataset is presented in Table 1.

Table 1 Erythemato-squamous diseases dataset from UCI Repository

| Classes of Diagnosis | Features (F) & Values | |
|---|---|---|
| | Clinical | Histopathological |
| $C_1$ : **psoriasis** | $f_1$ : erythema | $f_{12}$ : melanin incontinence |
| $C_2$ : **seboreic dermatitis** | $f_2$ : scaling | $f_{13}$ : eosinophils in the infiltrate |
| $C_3$ : **lichen planus** | $f_3$ : definite borders | $f_{14}$ : PNL infiltrate |
| $C_4$ : **pityriasis rosea** | $f_4$ : itching | $f_{15}$ : fibrosis of the papillary dermis |
| $C_5$ : **chronic dermatitis** | $f_5$ : koebner phenomenon | $f_{16}$ : exocytosis |
| $C_6$ : **pityriasis rubra pilaris** | $f_6$ : polygonal papules | $f_{17}$ : acanthosis |
| | $f_7$ : follicular papules | $f_{18}$ : hyperkeratosis |
| | $f_8$ : oral mucosal involvement | $f_{19}$ : parakeratosis |
| | $f_9$ : knee and elbow involvement | $f_{20}$ : clubbing of the rete ridges |
| | $f_{10}$ : scalp involvement | $f_{21}$ : elongation of the rete ridges |
| | $f_{11}$ : family history (0 = no,1= yes) | $f_{22}$ : thinning of the suprapapillary epidermis |
| | $f_{34}$ : age (linear) | $f_{23}$ : spongiform pustule |
| | | $f_{24}$ : munro microabcess |
| | | $f_{25}$ : focal hypergranulosis |
| | | $f_{26}$ : disappearance of the granular layer |
| | | $f_{27}$ : vacuolisation and damage of basal layer |
| | | $f_{28}$ : spongiosis |
| | | $f_{29}$ : saw-tooth appearance of retes |
| | | $f_{30}$ : follicular horn plug |
| | | $f_{31}$ : perifollicular parakeratosis |
| | | $f_{31}$ : inflammatory monoclear infiltrate |
| | | $f_{33}$ : band-like infiltrate |
| **number of Instances** | | 366 |
| **number of Attributes** | | 34 |
| **Missing attributes values (in age attribute) distinguished with ?** | | 8 |

**Table 2** Performance Evaluation Comparisons with respect to 10-fold Cross-validation

| Datasets | Performance Metrics | Naïve Bayes | Multilayer Perceptron | J48 |
|---|---|---|---|---|
| Erythemato-Squamous Disease (ESD) Dataset | Correctly Classified Instances | 97.4 | 96.6 | 93.5 |
| | Mean absolute error | 1.0 | 1.3 | 2.9 |
| | Root mean squared error | 7.4 | 9.7 | 13.9 |
| | Relative absolute error | 4.0 | 5.2 | 10.6 |
| | Root relative squared error | 20.4 | 26.7 | 38.2 |
| | True Positive (TP) Rate | 97.5 | 96.6 | 93.6 |
| | False Positive (FP) Rate | 0.4 | 0.7 | 1.9 |
| | Precision | 97.6 | 96.7 | 93.6 |
| | Recall | 97.5 | 96.6 | 93.6 |
| | F-Measure | 97.5 | 96.7 | 93.5 |
| | ROC Area (AUC) | 99.9 | 99.8 | 96.6 |

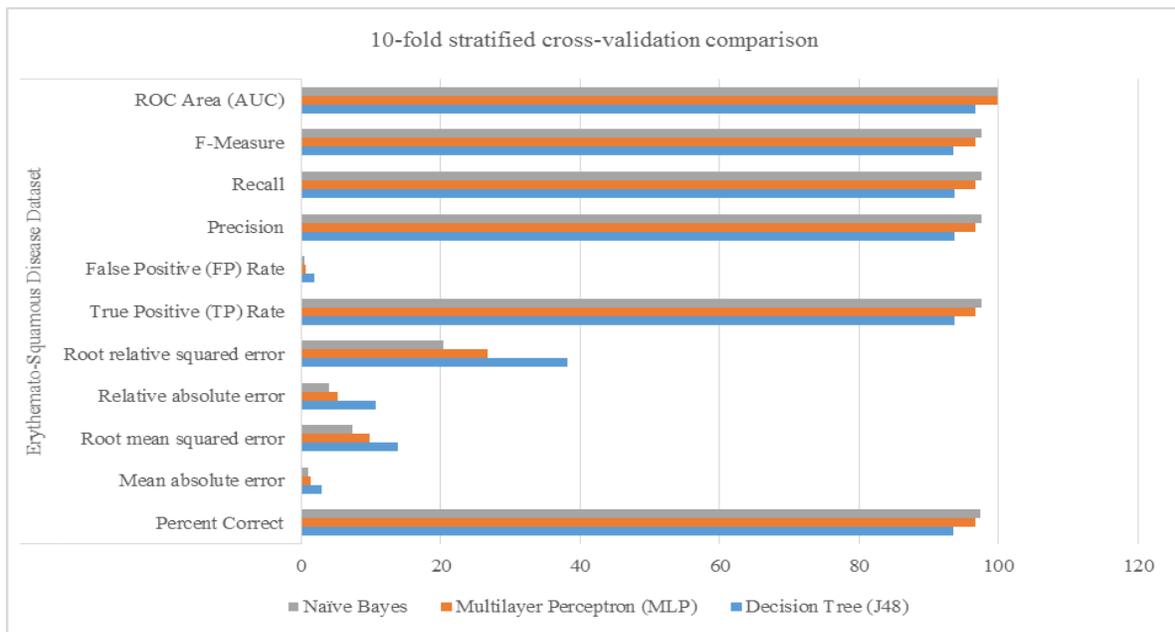

**Figure 3** 10-fold Stratified cross-validation performance comparison for ESD

## 6. Results and Discussion

We used Waikato Environment for Knowledge Analysis (WEKA) to simulate the baseline performance accuracy of the classifiers in a more convenient manner [48], to determine which scheme is statistically better than the others. We used stratified 10-fold cross-validation evaluation model in the final analysis. Stratified 10-fold cross-validation (k = 10) is the most common [49], and universal [50] evaluation models, with lower sample distribution variance compared to the hold-out cross-validation. In the final comparative analysis, the results shows that J48 classifier achieved classification accuracy of 93.5%, with a True Positive rate of 93.6% and a ROC Area (AUC) of 96.6%. The Multilayer Perceptron classifier achieved a classification accuracy and True Positive rate of 96.6%, and a ROC Area (AUC) of 99.8%. However, the Naive Bayes was found to have performed better J48 and Multilayer Perceptron classifiers. The Naïve Bayes achieved a classification accuracy of 97.4% with a True Positive rate of 97.5%, and a ROC Area (AUC) of 99.9%. Table 2 and Fig. 3 presents the complete comparative analysis in both tabular and graph format for better visualization and analysis.

## 7. Conclusions

In this paper, we set out to identify the most performing predictive data mining algorithms applied in the diagnosis of Erythemato-squamous disease (ESD). The ESD dataset was obtained from the University of California Irvine (UCI) machine learning repository database. We pre-processed the data to remove noise and smoothen the sample distribution to avoid bias analysis. We removed the missing values in the age attributes because it was difficult to specify the likely range of values to replace the missing values distinguished with '?', in Table without bias [30]. In this study, the family history attribute with value 1 means the symptom or feature is observed in the family and 0 otherwise. In other histopathological features, 0 means the absence of symptoms, 3 means the largest amount possible, and 1,2 means the relative intermediate values observed [29].

We used the Naive Bayes, Multilayer Perceptron and J48 decision tree predictive data mining algorithm to determine the ESD diseases. The three classifiers were calibrated to establish optimal parameter settings that maximizes the baseline performance accuracy of each classifier. Also, stratified 10-fold cross-validation evaluation model was used to measure the unbiased predictive accuracy of the three algorithms. We used 9 of 10 folds for training and the 10th fold for testing to ensure we obtained a less biased baseline performance accuracy measures to compare the three classifiers. The final comparative analysis of the models shows that the Naive Bayes classifier performed best with a classification accuracy of 97.4% with a True Positive rate of 97.5% and a ROC Area (AUC) of 99.9%. The Multilayer Perceptron classifier came out to be second best with a classification accuracy and True Positive rate of 96.6%, and a ROC Area (AUC) of 99.8%. And the J48 decision tree classifier came out to be the worst with classification accuracy of 93.5%, with a True Positive rate of 93.6% and a ROC Area (AUC) of 96.6%. In our future research, we hope to investigate the predictive abilities of Naive Bayes, Multilayer Perceptron and J48 decision tree induction in clinical prognosis. Sometimes clinical data are noisy and imbalanced. The right sampling techniques and the most appropriate evaluations metrics should be used to evaluate the performance of the classifiers, and subject such result to rigorous assessment before applying it to real-life clinical decision.